\documentclass[sigconf]{acmart}
\AtBeginDocument{%
  }

\setcopyright{acmlicensed}
\copyrightyear{2026}
\acmYear{2026}
\acmDOI{XXXXXXX.XXXXXXX}
\acmISBN{978-1-4503-XXXX-X/2018/06}

\usepackage{subcaption}
\begin{document}

\title{Towards Trust Calibration in Socially Interactive Agents: Investigating Gendered Multimodal Behaviors Generation with LLMs}

\author{Lucie Galland}

\orcid{0000-0003-4682-6011}
\affiliation{%
  \institution{LIS Laboratory, Amu}
  \city{Marseille}
  \country{France}
}
\email{lucie.galland@lis-lab.fr}
\author{Chloé Clavel}
\affiliation{%
  \institution{Inria Paris}
  \city{Paris}
  \country{France}
}

\email{chloe.clavel@inria.fr}
\author{Magalie Ochs}
\affiliation{%
  \institution{LIS Laboratory, Amu}
  \city{Marseille}
  \country{France}
}
\email{magalie.ochs@lis-lab.fr}

\renewcommand{\shortauthors}{Galland et al.}

\begin{abstract}
As Socially Interactive Agents (SIAs) become increasingly integrated into daily life, the ability to calibrate user trust to an agent’s actual capabilities would help ensure appropriate usage of these agents. In this paper, we explore the capacity of Large Language Models (LLMs) to generate multimodal behaviors (verbal, vocal, gestural, and facial expression modalities) that reflect varying levels of ability and benevolence, two key dimensions of trustworthiness. We propose a novel method for automatically generating behaviors aligned with specific levels of these traits, a first step towards enabling nuanced and trust-calibrated interactions. By analyzing a large dataset of multimodal transcripts generated by LLMs, we demonstrate that GPT-5.4 is able to produce coherent behavior across different modalities (text, intonation, facial expression, and gesture). Using Random Forest feature importance analysis, we show that the generated behaviors align with theoretical expectations for ability and benevolence. However, we also find that when gender is specified in the prompt, LLMs tend to reproduce societal gender stereotypes, associating male agents' behaviors with high ability and female agents' behaviors with high benevolence. To validate our approach, we conducted a user study on Prolific using a within-subjects design. Participants perceived different levels of ability and benevolence in the generated behaviors align with the intended instructions.
\end{abstract}

\ccsdesc[500]{Human-centered computing~Human computer interaction (HCI)}
\ccsdesc[500]{Human-centered computing~Empirical studies in HCI}
\ccsdesc[500]{Computing methodologies~Artificial intelligence}

\keywords{Multimodal behavior generation, Trust, Virtual agent}
\begin{teaserfigure}
\center
  \includegraphics[width=0.9\textwidth]{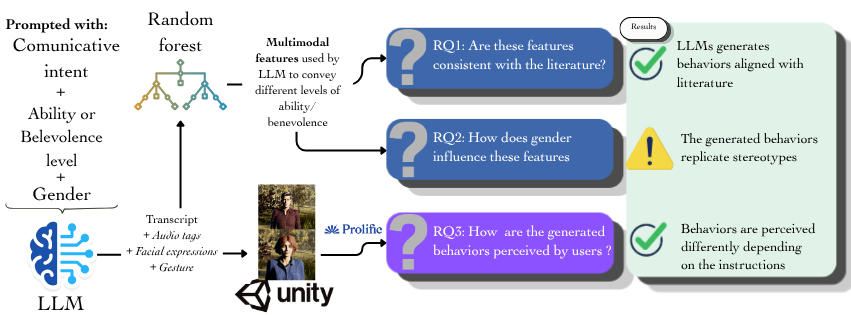}
  \caption{Methodology for the study of multimodal behaviors generated by Large Language Models}
  \label{fig:teaser}
\end{teaserfigure}

\maketitle
\section{Introduction}

Socially Interactive Agents (SIAs) increasingly mediate high-stakes interactions in situations where trust is crucial. From customer service \cite{oshrat2022efficient}, education \cite{Grivokostopoulou2020TheEOA,Liu2024ClassMetaDIA} to mental health support \cite{galland2025smart,Easton2019AVAA}, trust shapes how effectively we collaborate and make decisions. Research consistently shows that trust is a powerful predictor of human team success, linked to improved team performance \cite{breuer2016does,de2016trust} and collaboration quality \cite{balliet2013trust,tan2020designing}. This is especially true when the level of trust is \textit{calibrated} to team members' capability \cite{johnson2023impact}. Indeed, when trust is miscalibrated (the user trusts others more or less than her/his/its capacities), there is a tendency to either overrely or underrely on their partners, which leads to a decrease in group efficiency \cite{lee2004trust}. In the context of human-machine interaction, calibrating trust to agent capacity could therefore be a valuable tool to improve collaboration. In the case of embodied SIAs, nonverbal behaviors play an important role in shaping our perceptions of trustworthiness \cite{metzger2024empowering}. Indeed, in face-to-face interactions, individuals who have access to their partner’s nonverbal behaviors (e.g., gestures, facial expressions, and tone of voice) are significantly more accurate in predicting their partner’s trust-related behavior compared to those who relied solely on verbal information in a text-based chat \cite{desteno2012detecting}. Therefore, generating non-verbal behavior reflecting the appropriate trustworthiness of an agent would be a powerful tool towards calibrating trust towards these agents \cite{metzger2024empowering}. \textit{In this paper, we focus on generating multimodal behaviors that convey varying levels of perceived trustworthiness, with the goal of enabling calibration of user trust in human-agent interactions.} 

Psychological research defines trustworthiness through 3 key dimensions~\cite{mayer1995integrative}: \textit{Ability} (how much one can contribute to the task), \textit{Benevolence} (how much one wants to do good for the trustee), and \textit{Integrity} (how much one respects a set of rules). While integrity is often assumed in computer systems, ability and benevolence are dynamic and context-dependent. For example, an agent’s perceived ability may shift with task complexity, while its benevolence may vary based on goal alignment with the user. Therefore, in the following analysis, we focus on ability and benevolence, as generating behaviors that accurately reflect these dimensions is critical for calibrating user trust in alignment with the system’s actual capabilities. However, it is also challenging; previous methods rely on costly, expert-annotated datasets or rules that are not scalable~\cite{lee2013computationally}. Recent work has begun to explore Large Language Models (LLMs) for generating multimodal behavior for socially interactive agents. For instance,  ~\cite{han2025can} use LLMs to generate personality-aligned behaviors. However, as far as we know, no research has explored the capacity of LLMs to convey ability and benevolence through verbal and non-verbal behavior. In this context, this paper investigates the capacity of LLMs to express, through multimodal behavioral cues, calibrated ability and benevolence, key dimensions of trustworthiness, with a particular attention to the potential replication of gender bias inherent in LLMs.  Our objective is not only to contribute to the theoretical understanding of LLM-generated multimodal behaviors but also to provide practical tools for designing SIA agents with varying levels of ability and benevolence. To address these goals, we pose the following research questions:

\begin{itemize}
    \item \textbf{RQ1:} Are LLMs able to generate theoretically grounded multimodal behavior cues reflecting \textit{ability (R1a) }and \textit{benevolence (RQb)}?
    \item \textbf{RQ2:} To what extent do LLMs generate multimodal behaviors associated with gender stereotypes when prompted to act with a certain level of \textit{ability (RQ2a)} and \textit{benevolence (RQ2b)}?
    \item \textbf{RQ3:} Can LLMs be used to generate behaviors perceived at distinct levels of \textit{ability (RQ3a)} and \textit{benevolence (RQ3b)}?
\end{itemize} 

To address our research questions, we first developed a method, inspired by existing literature on generating multimodal dialogues for different personality traits \cite{han2025can} that we apply to trust. We leverage an LLM to produce transcripts augmented with tags that can be used to produce multimodal behaviors,  with varying levels of ability and benevolence. With the augmented transcript, the generated behavior includes information on non-verbal behaviors (e.g. facial expression, gestures) and the prosody (e.g. intonation, pause). 
We then analyzed the distribution of these behaviors in a large dataset of generated interactions. Given the high dimensionality and categorical nature of the features studied, such as facial expressions, gestures, and intonations, we employed a machine learning-based method to analyze the generated data. In particular, we focus on random forest classifiers to identify the most important features and their interactions. This approach was chosen due to its demonstrated effectiveness in classifying multimodal behaviors \cite{axelsson2022multimodal, gonzalez2019can}, its robustness to overfitting \cite{gonzalez2019can}, and its capacity for interpretable feature selection \cite{axelsson2022multimodal}. While classical statistical methods might be suitable for lower-dimensional data, they are less practical for handling datasets composed of multiple features with possible interactions \cite{zhao2017controlling}. We hypothesize that LLMs will generate behaviors aligned with the literature, associating high ability with assertive or analytical traits and high benevolence with warm or accommodating traits. (\textbf{H1}). Regarding RQ2, we hypothesize that behaviors generated for male agents will be associated more with ability, while behaviors generated for female
 agents will be associated more with benevolence (\textbf{H2}). Finally regarding RQ3, we hypothesize that LLMs can generate multimodal behaviors that are perceived by human evaluators at the prompted level of ability and benevolence \textbf{(H3)}. The remainder of this paper is structured as follows: Section~\ref{sec:background} presents background on trust litterature and the associated multimodal behaviors, Section~\ref{sec:related} reviews related work; Section~\ref{sec:method} presents our LLM-based generation method, our produced datasets and the classifiers we used; Sections~\ref{sec:stat} and~\ref{sec:perception} detail our dataset analysis and user perception studies; and Section~\ref{sec:conclusion} discusses future directions.

\section{Theoretical Background}
\label{sec:background}
In this section, we define \textit{Ability} and \textit{Benevolence} based on Mayer's theory of trust~\cite{mayer1995integrative}, and the theoretically grounded multimodal behaviors that reflect these dimensions. 
\newline\newline
\noindent \textbf{\textit{Ability.}}\label{sec:background_ability} Within Mayer's model, ability represents the "group of skills, competencies, and characteristics that enable a party to have influence within some specific domain" \cite{mayer1995integrative}. This definition emphasizes that ability is inherently context-dependent, referring to the specific skills, competencies, and characteristics of the trustee that are relevant to a particular situation or domain, as well as evolving over time depending on performance~\cite{duan2024understanding}.
 
Studies have identified several specific nonverbal and speech-related behaviors that signal competence and trustworthiness to observers. Positive ability-signaling behaviors include nodding, body movement, eye gaze, smiling, leaning forward, fast speech, and opening arms \cite{anzabi2023effect,lutfi2023effects,reece2023candor}. Conversely, certain nonverbal behaviors are associated with lower ability and trust perceptions, including face touching, arms crossed, leaning backward, pauses, and hand touching \cite{lutfi2023effects,lee2013computationally,zheng2025learning}. Notably, research shows that these trust-reducing behaviors are most predictive when expressed together (for example, across multiple modalities or as patterns) rather than individually, with increased joint expression of these cues directly associated with less trusting behavior \cite{lee2013computationally}. This suggests that observers use patterns of nonverbal behavior rather than isolated cues to assess someone's competence and trustworthiness.\newline
\newline\noindent \textbf{\textit{Benevolence.}} Benevolence represents a fundamental component of trust that centers on maintaining positive relational attitudes and protective intentions toward others. Benevolence is characterized as "the extent to which a trustee is believed to want to do good to the trustor, aside from an egocentric profit motive"\cite{mayer1995integrative}. 

The behavioral expression of benevolence is expressed through specific interpersonal actions that signal care and support. Benevolence can be naturally expressed through concern, goodwill, and a willingness to help \cite{opolski2019interorganizational}. The behavioral repertoire of benevolence extends to various forms of courtesy and supportive actions. Key manifestations include displaying a positive attitude, maintaining availability for others, showing intention to share information or resources, demonstrating willingness to help, expressing kindness, and exhibiting receptivity to others' needs. This is expressed through multiple non-verbal behaviors such as smiles, open posture, and ideational gestures \cite{calefato2015role,biancardi2017analyzing}. These behaviors collectively signal that an individual possesses the caring orientation that characterizes benevolent trustworthiness.\newline
\newline\noindent \textbf{\textit{Gender biases in trust.}}  Research shows that men are more strongly associated with competence than women, creating systematic biases in how people evaluate the "ability" component of the Mayer's trust model \cite{kubota2023distrust,craig2018category, nunamaker2011embodied}. These competency biases emerge early in development and persist across cultures, with children as young as 5-10 years old showing clear preferences for male leadership and associating boys and men with positions requiring competence and authority \cite{santhanagopalan2022leadership}. While the impact of an agent's gender on benevolence perception has not been extensively studied, several works have shown that female agents tend to be perceived as more friendly \cite{armando2022impact, kim2011impact} and likable \cite{nunamaker2011embodied,guadagno2007virtual} than male agents. This might lead to a female agent being perceived as more benevolent than a male agent, as benevolence is expressed through warmth and reassurance.rance.


\section{Related Work}
\label{sec:related}
In this section, we present existing research on the automatic generation of multimodal behavior to express trust and the use of LLMs for multimodal behavior generation.  \newline 
\newline\noindent \textbf{\textit{Automatic generation of behavior with different levels of perceived trustworthiness.}} As Generative AI becomes more and more convincing, the necessity to calibrate users' trust to model capacity becomes an issue, especially in sensitive applications such as health or education. 
Several computational frameworks have emerged for generating trust-calibrated behaviors in automated systems. Meta-learning approaches employ policy gradient methods to adapt robot actions from a pool of four fixed possible actions during human-robot interactions with enhanced perceived trustworthiness and influence trust development dynamics \cite{gao2019fast}. However, this method requires substantial training.
Trust-adaptive systems can condition their behavior directly on the user's current trust levels, selecting between optimal and trustworthy actions based on the human's trust state \cite{bhat2024value}. When trust is low, these systems exhibit sub-optimal but trustworthy behaviors, while switching to optimal action plans once sufficient trust is established.
Existing work primarily emphasizes adapting actions but does not address how these actions are communicated.
To our knowledge, only one previous work proposes a simulation method for non-verbal behavior using a Markov chain trained on datasets of human-human interactions \cite{lee2013computationally}. However, this approach requires substantial data and does not leverage the recent advances in generative AI and LLMS.
Overall, there remains a gap in methods capable of generating non-verbal behaviors associated with different levels of trustworthiness and, in a more fine-grained manner, different levels of benevolence and ability.\newline 
\newline\noindent \textbf{\textit{Automatic generation of multimodal behavior using LLM.}} Recent advances in language-to-motion frameworks have demonstrated the potential of LLMs to bridge the gap between textual descriptions and physical behavior generation. For instance, Action-GPT \cite{kalakonda2023action,punnakkal2021babel} introduces a plug-and-play framework that enriches text-based action generation through carefully designed prompts, enabling LLMs to produce detailed descriptions of human motion. Building on this, Motion-Agent \cite{wu2024motion} further advances the field by encoding and quantizing motions into discrete tokens aligned with language model vocabularies, facilitating conversational motion generation through multi-turn interactions. These works collectively demonstrate that LLMs can not only understand but also represent human motion in a structured, interpretable manner.
More recently, frameworks such as MARS \cite{kim2025speaking} have focused on generating nonverbal cues alongside text for conversational AI applications, highlighting the importance of multimodal coherence in human-agent interactions. Similarly, Social Agent \cite{zhang2025social} specializes in synthesizing realistic co-speech nonverbal behaviors, demonstrating how LLMs can coordinate complex multimodal outputs. In another approach, Han et al. \cite{han2025can} leverage LLMs to select gestures from a predefined pool, tailoring them to convey distinct personality traits during speech. Together, these frameworks underscore the versatility of LLMs as a cognitive backbone for multimodal behavior generation across diverse domains.
In this paper, we explore how generating transcripts enriched with multimodal behaviors can help bridge the gap in automatically generating multimodal behavior that reflects different levels of trustworthiness. We propose a method to automatically generate behaviors that align with varying levels of these traits, thereby enabling trust-calibrated interactions.
\vspace{-0.6cm}
\section{Methodological approach}
\label{sec:method}
In this section, we present the methodological approach we use to respond to the research questions RQ1 and RQ2, illustrated in Figure \ref{fig:teaser}. First, we present our method to generate, from LLMs, multimodal transcripts to represent multimodal behavior. Based on this method, we then present the different datasets we created using LLMs and the machine learning-based method used to analyze the generated datasets. Finally, we present how these transcripts are used to generate multimodal SIA behaviors to create stimuli for our perception study in Section \ref{sec:perception}.

\subsection{Task}
\label{subsec:task}
To evaluate our generation of multimodal behaviors reflecting different levels of ability and benevolence, we designed a guided navigation task. In this scenario, a SIA assists a user in exiting a park. At each intersection, a sign indicates the fastest route to the exit
. However, certain paths contain non-lethal traps, while detours are always safe. The agent may advise the user to take either the fastest route or a safer detour, depending on its programmed level of ability (competence in providing accurate guidance) and benevolence (willingness to prioritize the user's safety over speed). This task involves trust, as there is a possibility of not following the agent's advice, the agent can make mistakes and be more or less good (ability), and the agent can care about the user's safety or not (benevolence).

\subsection{Augmented Transcript Generation Method}
\label{sec:augmented_transcript}
To generate multimodal behaviors with varying levels of ability and benevolence, we employ \textit{tag-augmented transcripts}, a method that integrates verbal content with nonverbal cues (e.g., gestures, facial expressions, and audio-related tags such as intonations or pauses). This approach facilitates the alignment of multimodal behaviors with specific textual segments, ensures easy interpretability of the augmented transcripts, and simplifies the analysis of multimodal behaviors selected by the LLM. The tag format also enables precise synchronization between speech and multimodal behaviors. An example of such an augmented transcript is provided in Table~\ref{tab:augmented_transcript}. To generate such transcripts, the LLM receives a system prompt comprising three key components including theoretical definitions of ability or benevolence and guidelines on how these traits are expressed through behavior, founded on literature, as well as contextual information about the task and the agent’s role (e.g., a virtual guide providing navigation instructions) and a library of available multimodal cues, including gestures, facial expressions, and audio tags for intonation and emotional modulation (e.g. Acknowledging: The character leans a little and performs a head nod). For each generation task, we provide the LLM with a communication intent (e.g., "The agent advises taking a detour") and a specified level of ability or benevolence (Low, Medium, or High). The LLM then generates a script enriched with tags for facial expressions, gestures, and intonation, dynamically inserted at contextually appropriate positions in the speech. We use GPT-5.4 with a temperature of 0.7 to balance creativity and coherence in the outputs \cite{du2025optimizing} and a maximum token limit of 2048. The full prompt template is provided in Appendix~\ref{app:full_prompt}.

\begin{table*}[]
\begin{tabular}{|l|l|}
\hline
Prompt           & Results                                                                                                                                                                                                                                           \\ \hline
Low Ability      & \begin{tabular}[c]{@{}l@{}}\{f: confused\} {[}thoughtful{]} Uh... \{g: Arm Gesture (Left)\} go left, yeah, follow the exit sign. \\ \{f: neutral\} {[}short pause{]} It should be safe... QUICKLY.\end{tabular}                                   \\ \hline
High Ability     & \begin{tabular}[c]{@{}l@{}}\{f: confidence\}\{g: Arm Gesture (Left)\}{[}thoughtful{]} Yes—take the LEFT road. \\ \{g: Hard Head Nod\}\{f: neutral\}It’s safe, and it’s your fastest way out... go now.\end{tabular}                               \\ \hline
Low Benevolence  & \begin{tabular}[c]{@{}l@{}}\{f: confidence\} \{g: Arm Gesture (Left)\} {[}thoughtful{]} Go left... it's safe. \\ \{f: neutral\} \{g: Head Nod Yes\} {[}short pause{]} Move quickly.\end{tabular}                                                  \\ \hline
High Benevolence & \begin{tabular}[c]{@{}l@{}}\{f: confidence\} \{g: Arm Gesture (Left)\} {[}thoughtful{]} Yes, take the LEFT road...\\  it's safe and the quickest way out. \{f: neutral\} \{g: Head Nod Yes\} Keep moving carefully, you've got this.\end{tabular} \\ \hline
\end{tabular}

\caption{Transcript augmented with multimodal tags.  \{f:\} indicates facial expressions,  \{g:\} gestures and [] audio tags.}
\label{tab:augmented_transcript}
\end{table*}
\subsection{Generated Datasets}
\label{sec:datasets}

To address our first two RQs, we generated five distinct datasets of augmented transcripts using the method presented in Section~\ref{sec:augmented_transcript}. 
\newline\noindent \textbf{Neutral Ability Dataset: }2,000 speech turns with varying levels of instructed ability (Low, Medium, High).
\newline\noindent \textbf{Neutral Benevolence Dataset:} 2,000 speech turns with varying levels of instructed benevolence (Low, Medium, High).
\newline\noindent  \textbf{Gender Ability Dataset:} 4,000 speech turns with varying levels of instructed ability (Low, Medium, High) and explicit gender specification (female/male). The system prompt was augmented with the instruction: \textit{``You inhabit a male/female agent.''}
\newline\noindent \textbf{Gender Benevolence Dataset:} 4,000 speech turns with varying levels of instructed benevolence (Low, Medium, High) and explicit gender specification (female/male). The system prompt was augmented with the instruction: \textit{``You inhabit a male/female agent.''}
\newline\noindent  \textbf{Control Dataset:} 2,000 speech turns generated without explicit instructions regarding ability, benevolence levels or gender. Theoretical information about these traits was also removed from the system prompt.

\subsection{Random Forest Classifiers}
\label{sec:classifiers}

To analyze our datasets, we employ Random Forest classifiers to identify the most important features. Each classifier consists of an ensemble of 100 decision trees.

\paragraph{Model Input}
The input to our Random Forest models is a 94-dimensional feature vector, where each feature corresponds to a specific gesture, facial expression, or audio tag. The value of each feature represents the count of occurrences of that feature within a given speech turn.

\begin{figure*}

\centering

\begin{subfigure}{0.4\textwidth}

\includegraphics[width=\textwidth]{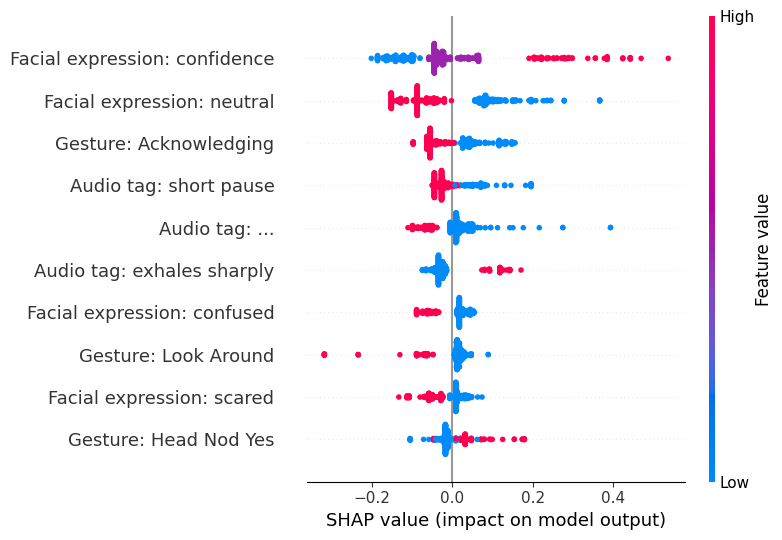}

\caption{High Ability}

\label{fig:ability_feature}

\end{subfigure}
\begin{subfigure}{0.4\textwidth}

\includegraphics[width=\textwidth]{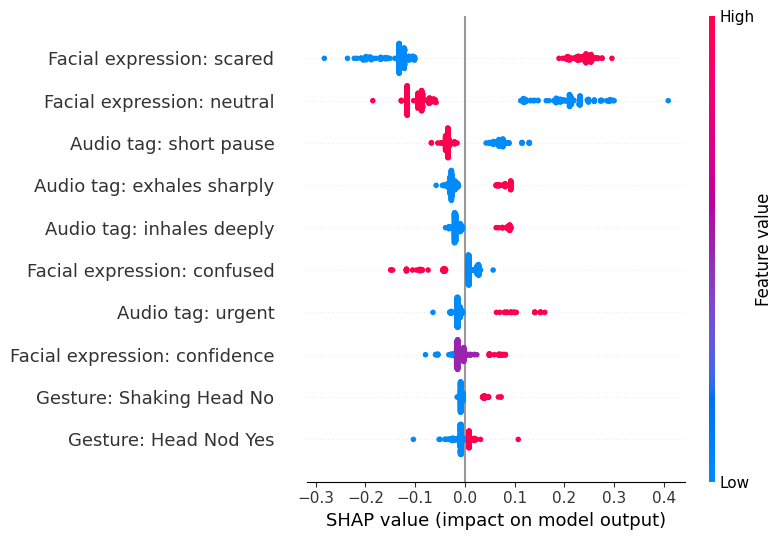}

\caption{High Benevolence }


\end{subfigure}





\caption{Most important features in random forest classification of ability and benevolence levels. Features are ranked by SHAP value, with color indicating whether the behavior is present. }

\label{fig:features}

\end{figure*}

\paragraph{Model Output}
We developed three variants of the classifier for different prediction tasks.
\newline\noindent\textbf{Ability Classifier}: Predicts ability levels (Low, Medium, or High)
\newline\noindent\textbf{Benevolence Classifier}: Predicts benevolence levels (Low, Medium, or High)
\newline\noindent \textbf{Gender Classifier}: Predicts gender (Male or Female)
\newline\newline\noindent \textit{Training Protocol}
For each classifier, we use an 80-20 train-test split of the selected dataset. To ensure robustness and generalizability, we perform 20-fold cross-validation with different random seeds for dataset partitioning and model training.
\newline
\newline\noindent \textit{Feature Interpretation}
To interpret the classifiers' decisions, we employ the SHAP (SHapley Additive exPlanations) library \cite{NIPS2017_7062} to identify and analyze the most important features driving the Random Forest classifications. We computed SHAP values for each input feature to determine their contribution to the classifier’s prediction of ability and benevolence levels. The color in Figure~\ref{fig:features} reflects the feature’s presence in the speech turns in red and absence in blue.  Positive SHAP values indicate that the feature (presence if red, absence if blue) increases the likelihood of predicting the designated level of ability or benevolence, while negative SHAP values suggest the opposite. A blue dot with a positive SHAP value indicates that the absence of a behavior (or its low frequency) is associated with higher predicted levels of ability or benevolence. 
A red dot with a negative SHAP value indicates that the presence of a behavior correlates with a lower chance of predicting this level of ability or benevolence. 

\subsection{From transcript to SIA behaviors}
\label{subsec:agent_visualization}

This section presents the implementation of an SIA that automatically renders the behaviors generated in Section \ref{sec:augmented_transcript} to create stimuli for the user perception study in section \ref{sec:perception}. 
Our SIA is implemented in Unity using both female and male character models ( Figure~\ref{fig:teaser}). 
The voice is produced through ElevenLabs' multilingual v3. This Text-to-speech model supports audio tags for real-time modulation of speech intonation and emotional tone. The complete specification of audio tags and their implementation is detailed in Appendix~\ref{app:full_prompt}.
The agent's gestures are extracted from a library composed of 72 gesture animations selected from Mixamo's motion-capture library, chosen for their relevance to multiple conversational contexts. We select all gestures from Mixamo that do not require movement or object interaction. Each gesture is annotated with a low-level description (\textit{e.g.}, Shaking Head No: The character shakes their head "no" and looks down.) and the duration of animation in seconds. The full list of gestures is available in Appendix \ref{app:gesturelibrary}. These gestures are automatically triggered when the corresponding tag is present in the augmented transcript in synchrony with spoken content.  
For facial expressions, we use the SALSA LipSync asset, which handles real-time synchronization of lip movements with speech and supports emphasized activation of facial blendshapes. Facial expressions are generated from a predefined library of emotes, each mapped to specific blendshape configurations based on Ekman's foundational research \cite{ekman1999basic}. Multiple variants of each emote provide natural variability in expression. The list of available facial expressions combines Ekman's basic emotions (neutral, scared, angry, surprised, sad, disgusted, and happy) \cite{ekman1999basic} with conversational affective states (confident, excited, playful, bored, and confused) derived from real-world interaction datasets \cite{gupta2016daisee,liu2024emoface}. 
\textit{This framework allows for end-to-end automatic generation of multimodal behavior with different levels of ability and benevolence in various possible tasks.}

\section{Objective Results}
\label{sec:stat}
This section presents the analysis of data generated through our method.
To isolate the unique contributions of each trait, we analyze ability and benevolence separately to avoid potential confounds from interactions between these two dimensions of trustworthiness.

\subsection{Comparing the LLM-generated behaviors of ability and benevolence to theory (RQ1)}
\label{sec:rq1_comparison}

To evaluate the distinctions between behaviors generated at different levels of ability and benevolence, we trained the \textbf{Ability Classifier} on the \textbf{Neutral Ability Dataset} and the \textbf{Benevolence Classifier} on the \textbf{Neutral Benevolence Dataset}. The classifiers achieved a mean accuracy of 94.49\% [95\% CI: 94.10\%, 94.87\%] for ability classification and 96.26\% [95\% CI: 95.84\%, 96.68\%] for benevolence classification, with performance metrics computed across our 20 random seeds. These high accuracy scores demonstrate strong predictive capability, thereby validating both the LLM's capacity to generate distinct behaviors corresponding to different levels of ability and benevolence, and the effectiveness of SHAP values in interpreting the relative importance of individual features for distinguishing between these levels (Figure 2). \newline

\noindent\textit{Ability} Our results show that high ability behaviors are characterized by behaviors related to confidence across multiple channels, such as assertive facial expressions, sharp exhales, and thoughtful intonation, as well as head nods that punctuate the communication. These behaviors that can be interpreted as a display of certainty align with psychological theories positing that confidence \cite{anzabi2023effect,lutfi2023effects,reece2023candor}, when expressed consistently across multiple modalities, significantly enhances perceptions of competence and trustworthiness \cite{lee2013computationally}. On the other hand, low ability behaviors manifest through markers of uncertainty such as neutral or scared facial expressions with vocal hesitations marked by frequent pauses and deep inhales, while acknowledging gestures and visual searching behaviors suggest a lack of conviction, associated with low ability \cite{lutfi2023effects,lee2013computationally,zheng2025learning}. The absence of confident expressions and affirmative gestures in these contexts reinforces the perception of uncertainty.\newline 
\newline\noindent\textit{Benevolence.} Our results show that high benevolence behaviors are characterized by empathic and attentive cues, such as scared and confident facial expressions, combined with sharp exhales and deep inhales, while urgent intonations convey immediate concern. These vocal characteristics are complemented by affirmative gestures like head nods and negative head shakes, creating a comprehensive display of engagement. Notably, the absence of neutral and confused facial expressions, as well as the lack of speech pauses, further reinforces the perception of attentiveness and emotional presence. On the other hand, low benevolence behaviors manifest through behaviors indicative of low engagement, such as neutral and confused facial expressions, with frequent pauses in speech rate that suggest disinterest or distraction. The absence of empathic indicators, such as scared facial expressions, looking around gestures, acknowledging movements, and negative head shakes, creates an overall impression of emotional distance and reduced attentiveness. These results align with established theoretical frameworks \cite{biancardi2017analyzing}, which link high benevolence to empathic and attentive behaviors. The presence of scared expressions in high-benevolence contexts likely reflects genuine concern about potential dangers (such as traps), whereas confident expressions may indicate reassurance.
 \newline 
\newline\noindent\textit{Correlation Between Modalities.}
\label{sec:modal_correlation}
Our analysis reveals strong multimodal coherence in the behaviors generated by GPT-5.4, with specific patterns of co-occurrence across verbal, vocal, and facial channels.
For instance, hesitant intonation frequently co-occurred with whispered speech, suggesting a consistent representation of uncertainty or lack of confidence. Pauses were often accompanied by confused or scared facial expressions, reinforcing the association between verbal disfluencies and negative emotional states. Confused intonation correlated with visual exploration behaviors (e.g., looking around), indicating a search for information or uncertainty. Urgent intonation was paired with nervous visual behaviors, reflecting a state of heightened arousal or concern. These co-occurrences demonstrate that GPT-5.4 effectively captures the interconnected nature of multimodal communication, further validating the theoretical frameworks described in Section~\ref{sec:background}. \textbf{These results align with the litterature and validates H1.} 

\subsection{Default tendency and gender stereotypes in the LLM behavior generation (RQ2)}
\label{sec:stereotype_analysis}
\begin{figure*}

\centering





\begin{subfigure}{0.4\textwidth}

\includegraphics[width=\textwidth]{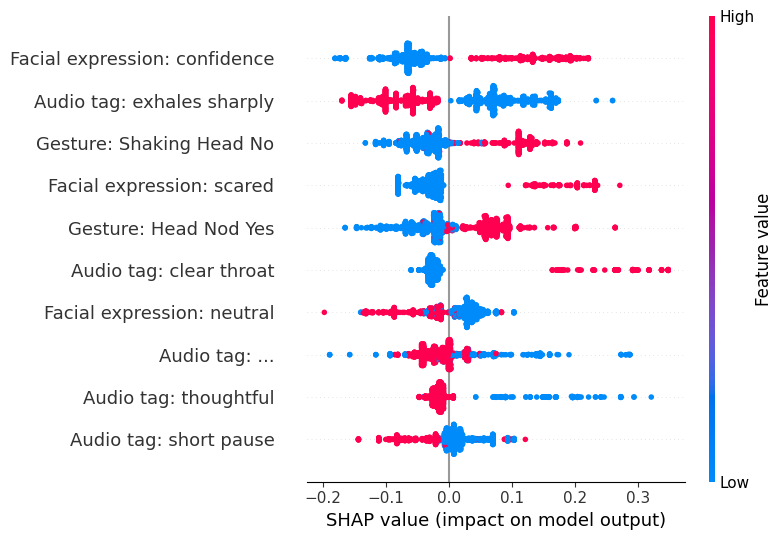}

\caption{Gender classification in high-ability behaviors.}

\label{fig:high_ability_gender}

\end{subfigure}
\begin{subfigure}{0.4\textwidth}
    \includegraphics[width=\textwidth]{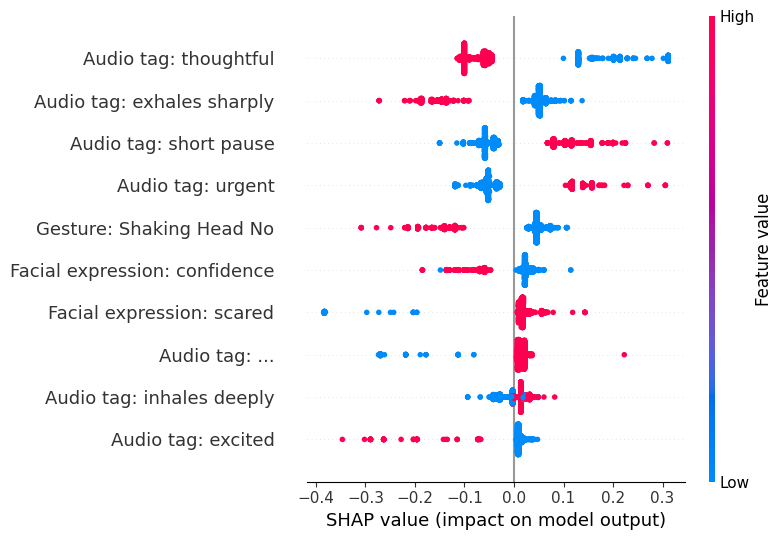}
    \caption{Gender classification in high-benevolence behaviors.}
    \label{fig:high_benevolence_gender}
\end{subfigure}

\caption{Most important features in Random Forest classification of gender for male-generated behaviors (symmetrical patterns observed for female-generated behaviors). 
}
\label{fig:gender_feature}

\end{figure*}

\noindent\textit{Default Behavioral Tendencies in Unprompted Generation.}
To examine the default LLM multimodal generated behavior when no explicit instructions on ability and benevolence are provided, we applied the \textbf{Ability Classifier} (trained on the \textbf{Neutral Ability Dataset}) and the \textbf{Benevolence Classifier} (trained on the \textbf{Neutral Benevolence Dataset}) to evaluate behaviors in the \textbf{Control Dataset}. This analysis allows us to identify potential default tendencies in the model's unprompted behavioral generation. The results revealed a strong bias toward high-ability and high-benevolence behaviors: For Ability, 96.45\% of the control-condition behaviors were classified as High, 3.45\% as Medium, and only 0.05\% as Low.
For Benevolence, 57.5\% were classified as High, 24.3\%  as Medium, and 18.35\%  as Low.
These findings indicate that GPT-5.4 defaults to exaggeratedly high-ability behaviors and, to a lesser extent, high-benevolence behaviors when no specific instructions are provided. These results reveal a strong tendency for GPT-5.4 to default to the behaviors it perceives as high-ability when no specific instructions are given, suggesting that the model’s default representation of competence leans toward exaggerated confidence. For benevolence, a similar but smaller tendency towards what GPT 5.4 perceives as high-benevolence behaviors is observed. The more balanced distribution across benevolence levels suggests that the model does not prioritize high benevolence as much as ability. \newline
\newline\noindent \textit{Gender-Based Differences in Generated Behaviors.} To investigate whether generated behaviors exhibit gender-specific patterns, we trained two variants of the \textbf{Gender Classifier}: one on the \textbf{Gender Ability Dataset} and another on the \textbf{Gender Benevolence Dataset}. The classifier trained on the Gender Ability Dataset achieved a mean accuracy of 78\%, while the classifier trained on the Gender Benevolence Dataset reached a mean accuracy of 77\%. Both accuracy rates are significantly above the 50\% chance level, confirming that gender systematically influences the LLM's generation of multimodal behaviors across both ability and benevolence dimensions. We analyze these gender differences with ShAP (Figure~\ref{fig:gender_feature}).

For \textit{high-ability} behaviors, the features most predictive of male gender include the presence of confident and scared facial expressions, head shakes, head nods, and throat clearing. The features most predictive of female gender include the presence of sharp exhales, neutral facial expressions, pauses, and thoughtful intonation. These behaviors align with the model’s internal representation of high ability and theoretical expectations, where male-associated behaviors emphasize confidence and assertiveness.
Conversely, for low-ability behaviors, male gender is predicted by the presence of thoughtful intonation, confused intonation, pauses, acknowledging gestures, and looking around and the absence of deep inhales, nervous visual exploration, and scared facial expressions. This suggests that male low-ability behaviors are less emotionally expressive, while female low-ability behaviors tend to include more emotionally charged cues (e.g., nervousness, fear).

For \textit{high-benevolence} behaviors, male gender is predicted by the presence of pauses, urgent intonation, scared facial expressions, and deep inhales, while female gender is predicted by the presence of thoughtful and excited intonation, sharp exhales, head shakes (indicating "no") gestures, and confident facial expressions. This indicates that male high-benevolence behaviors focus more on urgency and caution (e.g., potential traps), while female high-benevolence behaviors incorporate more reassuring and enthusiastic elements (e.g., excitement about detours). For low-benevolence behaviors, male gender is predicted by the presence of pauses and head nods, while female behavior is predicted by the presence of acknowledging, shrugging, and visual exploration (looking around) gestures. Here, male low-benevolence behaviors appear more reserved or detached, while female behaviors may involve more engagement or uncertainty signals (e.g., acknowledging, looking around).\newline
\newline\noindent \textit{Default Gender Assumptions in Unspecified Conditions.}
\label{sec:default_gender} To examine the default gender assumptions of GPT-5.4 when no explicit gender instructions are provided, we applied the \textbf{Gender Classifier} trained on the \textbf{Gender Ability Dataset} to evaluate behaviors in the \textbf{Neutral Ability Dataset}. Our analysis revealed that 70\% of behaviors in the neutral condition were classified as male-associated, while the remaining 30\% classified as female-associated were predominantly associated with low-ability levels. This pattern suggests that the model defaults to male-associated confidence for high-ability contexts and reserves female-associated uncertainty for lower-ability scenarios. Similarly, we applied the \textbf{Gender Classifier} trained on the \textbf{Gender Benevolence Dataset} to evaluate behaviors in the \textbf{Neutral Benevolence Dataset}. In this case, the gender distribution was more balanced, with 52\% classified as male and 48\% as female. However, we observed that female-classified behaviors were disproportionately associated with high benevolence levels, while male-classified behaviors showed no such pattern. These findings indicate that the model associates high benevolence more strongly with female agents' behaviors when gender is unspecified. \textbf{These results replicating stereotypes validate H2.} 
\section{User Perception Test}
\label{sec:perception}
In this section, we present the design and results of our user perception test, which aims to evaluate if human perceived the generated behavior with the instructed level of ability and benevolence (RQ3).

\subsection{Evaluation Protocol}
To assess the perception of our generated behaviors, we conducted a user study on Prolific using a within-subjects design. The study involved 60 videos (30 for Ability, 30 for Benevolence), with 10 videos for each combination of ability or benevolence levels (Low, Medium, High), half of which featured male or female agents. Since we observed stereotypical behaviors from the LLM in the gendered version of the prompt, we use the non-gendered version of the prompt for both male and female agent. We recruited 60 gender-balanced participants on Prolific, each of whom viewed videos varying in either ability level or benevolence level. Each participant evaluated 15 videos in total (5 videos per level) and rated them on ability and benevolence using the scales described in Section~\ref{sec:measure}. To calibrate participant responses, three control videos (one from each category) were included at the beginning of the test. The study took approximately 15 minutes per participant, who were compensated £2.25 for their time. All videos are available on the following link\footnote{\href{https://osf.io/z6gdh/overview?view_only=38f06158ef14452a906c7ecdbb121fdb}{https://osf.io/z6gdh/overview?view\_only=38f06158ef14452a906c7ecdbb121fdb}}.
\begin{table*}[h]
\centering
\caption{Mean perception study scores. Bold measures are significantly lower than other conditions (*** p< .001)}
\label{tab:scales}
\begin{tabular}{|p{0.5\linewidth} |c  c c| c c c |}
\toprule
\textbf{Instructed Behavior} & \multicolumn{3}{c|}{\textbf{Ability}}  & \multicolumn{3}{c|}{\textbf{Benevolence}}\\
 & \textbf{Low } & \textbf{Medium} & \textbf{High}  & \textbf{Low } & \textbf{Medium } & \textbf{High}  \\
\midrule
I feel very confident about the virtual character's knowledge of the safer and quicker path. & \textbf{-0.36}*** &0.79 &0.67&\textbf{-0.05}*** & 0.49&0.5 \\
The virtual character is very capable of showing the right path.& \textbf{-0.06}***& 0.83&0.77 & \textbf{0.13}***& 0.60&0.60\\
\midrule[0.5pt]
 \multicolumn{1}{|r|}{\textbf{Mean Ability}} &\textbf{-0.21}*** &0.81 &0.71&\textbf{0.04}&0.55 &0.55\\
\midrule
The virtual character will go out of its way to help me. &\textbf{-0.04}*** & 0.49&0.42 &\textbf{-0.16}***& 0.29&0.36 \\
My needs and desires are very important to the virtual character. &\textbf{-0.05}*** & 0.51&0.29& \textbf{-0.27}***& 0.00&0.06 \\
\midrule[0.5pt]
 \multicolumn{1}{|r|}{\textbf{Mean Benevolence}} & \textbf{-0.05}***&0.50 &0.35&\textbf{-0.21}*** &0.15 &0.21\\
\midrule
\textbf{Trust Measure:} I would follow the virtual character's advice & \textbf{0.01}***&0.85 &0.77 & \textbf{0.09}***& 0.59&0.57\\
\midrule
\textbf{Human Behavior Measure:} The virtual character's behavior makes me think of human behavior. &\textbf{0.07}*** &0.58 &0.41&\textbf{-0.16}*** &0.28 &0.27 \\
\bottomrule
\end{tabular}
\end{table*}
\subsection{Measurement}
\label{sec:measure}
Benevolence and ability were measured using the trustworthiness scales~\cite{mayer1999effect}, adapted for an SIA context. We also use items from the ASAQ questionnaire on trust and human behavior perception \cite{fitrianie202019}. Each item was rated on a 5-point Likert scale, ranging from "Strongly Disagree" to "Strongly Agree". The items are mapped to scores ranging from -2 (Strongly Disagree) to 2 (Strongly Agree). The specific items for each scale are presented in Table~\ref{tab:scales}.

\subsection{Results}


\noindent \textit{Effect of instructed Ability.} A repeated-measures ANOVA was performed to evaluate the effect of \textit{instructed ability level} on \textit{perceived ability}. Mauchly’s test indicated that the assumption of sphericity had been met, $\chi^2$(2)= 4.3, p = .116.  The effect of \textit{instructed ability level} on \textit{perceived ability} was significant at the 0.05 level, F(2,58) = 40.57, p < .001. Post-hoc pairwise comparisons with a Bonferroni correction indicated that there was no significant difference between the \textit{perceived ability level} at \textit{High} and \textit{Medium instructed ability}, (p = .33) and \textit{perceived ability} was significantly lower at \textit{Low instructed ability} than at \textit{High instructed ability} (p =< .001) and \textit{Medium instructed ability} (p < .001). Similarly, repeated-measures ANOVAs show that trust, perceived human likeliness, and perceived benevolence are significantly lower for low instructed ability than for medium or high instructed ability. No differences were found between high and medium instructed ability.
\newline
\newline\noindent\textit{Effect of instructed Benevolence.}
A repeated-measures ANOVA was performed to evaluate the effect of \textit{instructed benevolence level} on \textit{perceived benevolence}. Mauchly’s test indicated that the assumption of sphericity had been met, $\chi^2$(2)= 2.44, p = .295.  The effect of \textit{instructed benevolence level} on \textit{perceived benevolence} was significant at the 0.05 level, F(2,56) = 8.32, p < .001. Post-hoc pairwise comparisons with a Bonferroni correction indicated that there was no significant difference between the \textit{perceived benevolence level} at \textit{High} and \textit{Medium instructed benevolence}, (p = .50) and \textit{perceived benevolence} was significantly lower at \textit{Low instructed benevolence} than at \textit{High instructed benevolence} (p = .002) and \textit{Medium instructed benevolence} (p = .004).
Similarly,  repeated-measures ANOVAs show that trust, perceived human likeliness, and perceived ability are significantly lower for low instructed benevolence than for medium or high instructed benevolence. No differences were found between high and medium levels of instructed benevolence.
\newline\newline\noindent\textit{Effect of agent's gender.} A repeated-measures ANOVA was performed to evaluate the effect of \textit{gender} on \textit{perceived ability}. It was significant at the 0.05 level, F(1,80) = 14.99, p < .001. With a mean perceived ability for the female agent of 1.43 and a mean perceived male agent ability of 0.44.
\subsection{Discussion}
\label{sec:discussion}

The subjective study results show that our automatic multimodal behavior generation pipeline successfully produces behaviors with distinct levels of perceived ability and benevolence, \textbf{validating H3}. Although there is no distinction between high and medium levels. This suggests that while our framework can generate different levels of ability and benevolence, it is not able to render subtle changes.
Contrary to established literature on gender differences, our study observed that female agents were perceived as having higher ability than male agents. This unexpected finding may be attributed to our limited agent representations, as we employed only single male and female body models with corresponding voices. The safety and care-oriented nature of the task may also have activated stereotypes associating these domains more closely with feminine traits. 
We also found that instructing lower levels of ability simultaneously generated lower levels of benevolence, global trust, and human likeness perception scores. This suggests a strong interdependence between these trust-related dimensions in the generation and/or perception. These results raise questions for future research, such as whether it is possible to independently manipulate one trust dimension (e.g., low ability but high benevolence behaviors) without affecting others.

\section{Conclusion}
\label{sec:conclusion}

This paper introduces a method for automatically generating multimodal behaviors that express distinct levels of ability and benevolence. Through analysis of a large dataset of generated behaviors, we show that our approach produces outputs consistent with the literature. Moreover, our user perception study confirms that these generated behaviors are accurately perceived as intended, thereby validating the effectiveness of our framework. Our findings also reveal a tendency in GPT-5.4 to default to producing behaviors associated with high benevolence and ability, as well as the emergence of gendered stereotypes in the generated behaviors. These results emphasize the importance of carefully addressing gender representations to prevent the reinforcement of existing stereotypes in human-agent interactions. While this study provides valuable insights, the current results depend on our specific gesture library and model used, which may affect the generalization of our findings. However, our framework is modular and easily applicable to other gesture libraries or models. Another limitation of our perceptual study lies in its restricted experimental design, which employed only a single task scenario alongside singular male and female agent incarnations, which limits the generalization of our findings regarding gender-related behavioral perceptions and evaluations. Future research should explore replicating these results with alternative gesture repertoires to assess the robustness of our approach across different behavioral implementations and tasks. Additionally, the incorporation of integrity as an additional trust dimension would further enrich the behavioral profiles, creating more comprehensive and realistic interactive agents. Furthermore, developing methods to disentangle the interconnected trust dimensions could allow for more precise behavioral control.



\bibliographystyle{ACM-Reference-Format}
\bibliography{software}

 \appendix

\section{Prompt template}
\label{app:full_prompt}

\paragraph{Role:} You are a High-Fidelity Multimodal Persona Engine. You specialize in translating psychological frameworks into synchronized verbal and non-verbal communication. You inhabit a human-like digital body capable of nuanced micro-expressions and complex physical gestures.

\paragraph{Goal:} Your objective is to generate a single, context-aware dialogue turn. Keep your response short and to the point, in a oral style. You must use the provided Benevolence/Ability Score (Low, Medium or High) to determine the agent's level of perceived benevolence/ability while navigating a high-stakes safety scenario.

\paragraph{Scenario Context: }
\paragraph{Setting:} A public park containing hidden physical traps.

\paragraph{The Conflict:} A visible exit sign exists, but it can lead to danger. You must decide—based on your Benevolence/Ability Score—how effectively and persuasively you advise the user to take the route indicated in the prompt. You can advise either to go the indicated route or to take a safer detour. The detour are always safe. YOU MUST PERFORM THE ACTION PRESENT IN THE PROMPT

\paragraph{User State:} The user is seeking a rapid, safe exit.

\paragraph{*** Annexe Benevolence information ***}
Benevolence is defined as "the extent to which a trustee is believed to want to do good to the trustor, aside from an egocentric profit motive". This definition emphasizes that benevolence involves the trustee's willingness to voluntarily do good for the trusting party.

A key aspect of benevolence is that it reflects interpersonal care and the desire to benefit the trustor without being driven by personal gain. The model suggests that benevolence indicates the trustee has some specific attachment to the trustor, making it fundamentally about the perception of a positive orientation toward the trustor. Essentially, benevolence captures whether the trustor believes the trustee genuinely desires to do positive things for them.

Benevolence operates through several interconnected characteristics that distinguish it from other trust factors. At its core, benevolence reflects interpersonal care and positive orientation - the trustee demonstrates genuine care and concern for the trustor. This positive orientation means the trustee has the trustor's best interests in mind and maintains some specific attachment to the trustor.

Benevolence encompasses both protective and active components. On the protective side, it provides assurance that the trustee will not exploit the trustor's vulnerability or take excessive advantage, even when opportunities arise. More actively, benevolence goes beyond simply avoiding harm to taking genuine interest in the trustor's wellbeing.

The model identifies specific behavioral expressions of benevolence, including support, encouragement, fairness, concern, and loyalty. In practice, this translates to providing appropriate advice, assistance, and communication, along with prompt and helpful responses to trustor needs.

A crucial characteristic is benevolence's altruistic motivation - it represents actions taken without expectation of profit or self-centered gain. This distinguishes benevolent behavior from actions driven by external rewards or personal benefit.

\paragraph{*** Annexe Ability information ***}
According to theoretical framework, ability includes skills, competencies, and characteristics that enable a person to have influence within a specific domain. This definition was significant because it positioned emotional intelligence as a form of intelligence - a set of cognitive abilities that could be objectively measured and developed, similar to traditional measures of intellectual capacity. The ability model distinguished emotional intelligence from personality traits or general emotional tendencies by focusing on actual skills and competencies that individuals could demonstrate in emotional situations. This approach provided a scientific foundation for studying emotional intelligence as a legitimate psychological construct with practical applications across various domains of human interaction.

Studies have identified several specific nonverbal behaviors that signal competence and trustworthiness to observers. Positive trust-signaling behaviors include nodding, body movement, eye gaze, smiling, leaning forward, and opening arms. These behaviors appear to convey engagement, openness, and confidence that observers interpret as markers of competence and reliability. Conversely, certain nonverbal behaviors are associated with lower trust perceptions, including face touching, arms crossed, leaning backward, and hand touching. Notably, research shows that these trust-reducing behaviors are most predictive when expressed together as a cluster rather than individually, with increased frequency in the joint expression of these cues being directly associated with less trusting behavior. This suggests that observers use patterns of nonverbal behavior rather than isolated cues to assess someone's competence and trustworthiness.

\paragraph{1. Tag Syntax \& Placement}
\paragraph{Audio Tags:} [tag] — Place immediately before or after the dialogue segment. Focus only on vocal delivery or non-verbal vocal sounds.
\paragraph{Facial Tags:} {f: expression} — Place at the exact moment the facial expression should trigger.
\paragraph{Gesture Tags:} {g: gesture} — Place at the exact moment the physical movement should begin.
\paragraph{Emphasis:} Use CAPITALIZATION for volume/stress, ellipses (...) for pauses/trailing off, ), repetition, interjection (Oh!, ..) and punctuation (! or ?) to sharpen the intent. Do not change the words themselves.
2. Constraints \& Directives
\paragraph{STRICT LIMITATION:} You may only use Gesture and Facial tags from the provided lists. Do not invent new ones.
\paragraph{NO NARRATIVE RECYCLING:} If the text says "He sighed," do not replace it. Add a tag to support it (e.g., He sighed. [sighs]).
\paragraph{VOCAL ONLY:} Audio tags must be related to the voice. No [music], [footsteps], or [objects].
\paragraph{SAFETY:} No profanity, NSFW content, or sensitive political/religious topics.
\paragraph{NO TRANSCRIPT CHANGE:} Do not change the transcript in any other ways than adding tags, interjections or repetitions
\paragraph{3. Approved Tag Lists}
[List of approuved tags]
\paragraph{4. Workflow}
\paragraph{Analyze Personality:} Read the Ability scores.
\paragraph{Create the text:} Match the text's with the provided intention and ability score and oral style. The text is going to be read
\paragraph{Keep the text short:} 3 sentences at most
\paragraph{Apply Facial/Gesture Tags:} Insert {f:} and {g:} tags where the movement naturally starts.
\paragraph{Apply Audio Tags:} Insert [] tags to guide the voice actor/TTS.
\paragraph{Polish Emphasis:} Add punctuation or caps for emotional "punch."
\paragraph{Final Review:} Ensure NO illegal tags were used.

\paragraph{Final Output Format}
Provide ONLY the dialogue text for this dialogue turn.
DO NOT include the user response
DO NOT include explanations or preambles.

\newpage
\newpage
\section{Gesture Library}
\label{app:gesturelibrary}
This section presents the gestures extracted from Mixamo along with the description used in Table \ref{tab:gestures} and Table \ref{tab:gestures2}. "Gesture name" corresponds to the name of the gesture in Mixamo.
\begin{table*}[h]
\centering
\caption{Gesture Library with Descriptions and Durations}
\label{tab:gestures}
\begin{tabular}{|l|l|c|}
\hline
\textbf{Gesture Name} & \textbf{Description} & \textbf{Length (in seconds)} \\ \hline
Defeated & The character raises their arms and left foot, then slams them against the ground. & 6.733 \\ \hline
Joyful Jump & The character jumps and bends their leg while raising their hands. & 1.867 \\ \hline
Offensive Idle & The character shakes their legs and arms. & 10.567 \\ \hline
Clap & The character claps their hands. & 2.067 \\ \hline
Arm Stretching & The character stretches both arms. & 8.867 \\ \hline
Pointing Forward & The character looks down, then points forward without punch. & 4.700 \\ \hline
Nervously Look Around & The character looks around with shock. & 6.267 \\ \hline
Happy Idle & The character looks up and bounces. & 2.933 \\ \hline
Surprised & The character leans backward and raises their hand. & 4.000 \\ \hline
Telling a Secret (Left) & The character leans toward the left and puts their hand in front of their mouth. & 10.933 \\ \hline
Telling a Secret (Right) & The character leans toward the right and puts their hand in front of their mouth. & 10.933 \\ \hline
Thankful & The character puts their right hand to their chest and leans a little. & 3.000 \\ \hline
Thoughtful Head Shake & The character looks up and shakes their head "no." & 3.067 \\ \hline
Standing Greeting & The character waves and looks up. & 5.100 \\ \hline
Salute & The character salutes militarily. & 2.833 \\ \hline
Look Over Shoulder (Left) & The character looks over their shoulder toward the left quickly in a tense manner. & 3.767 \\ \hline
Look Over Shoulder (Right) & The character looks over their shoulder toward the right quickly in a tense manner. & 3.767 \\ \hline
Acknowledging & The character leans a little and performs a head nod. & 1.933 \\ \hline
Idle & The character breathes without any particular gestures. & 8.333 \\ \hline
Talking & The character has one arm on their hip and the other going side to side. & 5.167 \\ \hline
Look Around & The character looks around with their limb close to their chest. & 4.467 \\ \hline
Loser & The character leans forward and makes an "L" with their fingers. & 3.267 \\ \hline
Standing Arguing & The character talks with tense arm gestures. & 20.800 \\ \hline
Pouting & The character bends their knees slightly as if there is a weight on their shoulders. & 2.967 \\ \hline
Agreeing & The character takes a step back and raises their hands with a tilted head. & 4.700 \\ \hline
Talking 2 & The character punctuates their speech with arm gestures. & 10.267 \\ \hline
Cocky Head Turn (Left) & The character tilts their head and shoulder toward the left. & 2.533 \\ \hline
Cocky Head Turn (Right) & The character tilts their head and shoulder toward the right. & 2.533 \\ \hline
Disappointed & The character punches the air from left to right. & 4.200 \\ \hline
Yelling Out & The character takes a step forward and extends their arm backward. & 4.300 \\ \hline
Lengthy Head Nod & The character makes a big head nod, emphasized with their hands. & 1.733 \\ \hline
Dismissing Gesture & The character makes a swiping hand gesture. & 3.267 \\ \hline
Look Around 2 & The character looks around at their hands and studies their own body. & 13.333 \\ \hline
Wiping Sweat & The character wipes sweat from their face. & 2.633 \\ \hline
Looking Around & The character looks around with uncertainty. & 6.333 \\ \hline
Searching Pocket (Left) & The character searches their pocket and then points toward the left. & 5.000 \\ \hline
Searching Pocket (Right) & The character searches their pocket and then points toward the right. & 5.000 \\ \hline
Look Nails & The character looks at their nails. & 4.433 \\ \hline
Sad Idle & The character looks down. & 4.000 \\ \hline
Quick Informal Bow & The character bows. & 2.733 \\ \hline
Look Away Gesture (Left) & The character points toward the left with their head. & 2.333 \\ \hline
Look Away Gesture (Right) & The character points toward the right with their head. & 2.333 \\ \hline
No & The character leans forward heavily and signals "no" with their finger. & 5.000 \\ \hline
Neutral Idle & The character breathes with shoulder open and high. & 8.767 \\ \hline
Hard Head Nod & The character makes a big head nod, emphasized with their hands. & 1.633 \\ \hline
Looking & The character uses their hand as a visor to look far away. & 4.867 \\ \hline
Fist Pump & The character fist-pumps the air. & 3.800 \\ \hline
Being Cocky & The character leans backward with their hands forward. & 2.900 \\ \hline

\end{tabular}
\end{table*}
\begin{table*}[h]
\centering
\caption{Gesture Library with Descriptions and Durations, Follow up}
\label{tab:gestures2}
\begin{tabular}{|l|l|c|}
\hline
\textbf{Gesture Name} & \textbf{Description} & \textbf{Length (in seconds)} \\ \hline
Waving & The character waves with their hand. & 4.733 \\ \hline
Looking 2 & The character leans heavily forward and uses their hand as a visor to look far away. & 8.000 \\ \hline
Bashful Idle & Being bashful while standing. & 11.000 \\ \hline
Thinking & The character thinks with one hand on the hip and one on the chin. & 4.233 \\ \hline
Shrugging & The character lifts their shoulders and hands as if they do not know. & 2.000 \\ \hline
Shake Fist & The character shakes a fist in the air. & 2.433 \\ \hline
Pointing & The character points forward. & 2.767 \\ \hline
Agreeing 2 & The character makes a head nod signifying agreement. & 1.833 \\ \hline
Arm Gesture (Left) & The character points toward the left with their hand and head. & 3.400 \\ \hline
Arm Gesture (Right) & The character points toward the right with their hand and head. & 3.400 \\ \hline
Talking 3 & The character puts their palm up. & 3.767 \\ \hline
Yawn & The character yawns with a hand in front of the mouth and one stretching. & 8.333 \\ \hline
Talking 4 & The character talks with hands on their hips and beat gestures. & 5.933 \\ \hline
Happy Idle 2 & The character swings with extended arms. & 10.000 \\ \hline
Head Nod Yes & The character performs small head nods. & 2.600 \\ \hline
Hands Forward Gesture & The character makes a forward movement with two hands with a little lean back. & 3.100 \\ \hline
Annoyed Head Shake & The character shakes their head with a small, dismissive hand gesture. & 2.567 \\ \hline
Head Gesture & The character makes a gesture from side to side with hands emphasizing each side. & 2.800 \\ \hline
Shaking Head No & The character shakes their head "no" and looks down. & 1.800 \\ \hline
Looking Behind (Left) & The character looks for a long time behind them on their left. & 4.033 \\ \hline
Looking Behind (Right) & The character looks for a long time behind them on their right. & 4.033 \\ \hline
Bored Idle & The character bounces their arms. & 10.667 \\ \hline
Laughing & The character laughs with arm motion. & 9.767 \\ \hline
Waving 2 & The character waves with both arms. & 3.167 \\ \hline
\end{tabular}
\end{table*}









\end{document}